\documentclass[10pt,twocolumn]{article}
\usepackage[a4paper,margin=0.8in,columnsep=0.25in]{geometry}
\usepackage[T1]{fontenc}
\usepackage[utf8]{inputenc}
\usepackage{lmodern}
\usepackage{microtype}
\usepackage{amsmath,amssymb}
\usepackage{graphicx}
\usepackage{booktabs}
\usepackage{tabularx}
\usepackage{array}
\usepackage[font=small,labelfont=bf]{caption}
\usepackage[hidelinks]{hyperref}

\newcolumntype{Y}{>{\raggedright\arraybackslash}X}
\newcommand{\autofive}{auto@5}

\title{\Large\bfseries Zuse: A Decision Model for European Languages}
\author{Umair Mehmood Imam\\Founder and Builder, Bineric AI\\Assistant Professor, OsloMet, Norway\\\texttt{support@bineric.com}\\ORCID: \href{https://orcid.org/0009-0002-7892-9189}{0009-0002-7892-9189}}
\date{November 2026}

\begin{document}
\twocolumn[
\begin{@twocolumnfalse}
\maketitle
\begin{abstract}
\noindent Decision models read a document and return a probability for each of a fixed set of options instead of generating text. Several appeared in 2026, but none has published results outside English. We describe Zuse, three decision models (4B, 9B and 27B) that train a low-rank adapter and a pointer head on frozen open-weight bases, targeting 27 European languages, and use them to study how this design behaves across languages and tasks. On a broad European suite of 11 sources excluded from training, covering 9 task groups and 27 languages, a development set that also guided model selection, the 27B model reaches 87.5\% accuracy with a pooled expected calibration error of 0.016, and 0.035 when weighted per source. Post-hoc analyses of these single-seed results give three findings. First, miscalibration is structured by task rather than language: on parallel reading and topic items in 26 languages the confidence gap varies about as much as sampling noise, while across sources it spans $-0.07$ to $+0.09$, with knowledge tasks overconfident and acceptability judgements underconfident at all three sizes, so one fitted temperature cannot correct both. Second, the share of decisions that can be automated at a 5\% error budget follows chance-corrected skill (Spearman 0.97 over 33 source and model pairs) more closely than raw accuracy (0.91). Third, reach beyond the training languages is set by the base model: across 210 languages, changing the training run moved scores by at most about 2 points within a size, while languages outside the base's documented list trail by 23 to 28 points at every size. On English public suites Zuse is behind Jev and Kev, including Kev models on the same bases. We also report two pitfalls in building multi-source training data: label-word collision between topic schemes, and an injection augmentation that taught a shortcut. Zuse can be tried at \url{https://bineric.com/zuse}.
\end{abstract}
\vspace{1em}
\end{@twocolumnfalse}
]

\section{Introduction}

Many operational uses of language models require a choice among known options rather than generated text. Ticket routing, policy checks, moderation and form extraction all have this shape. A decision model suits such tasks: it reads a document and a set of questions, then returns a probability for each option. Because it cannot produce text, its output stays inside the options supplied, and its probabilities can be used to decide which cases to automate and which to send to a person.

This model class has developed quickly, with Jev~\cite{r1}, the open reproduction Kev~\cite{r2} and Clef~\cite{r3} all appearing in 2026. We found no per-language evaluation for any of them. Language coverage matters in Europe, where public bodies and companies work across many languages, including several with few speakers and limited training data. Whether the design transfers across such languages has not been measured in public.

We ask whether the decision-model design carries across European languages, and what governs calibration and automation there. We study this with Zuse, a family of three decision models targeting 27 European languages. We make six contributions:

\begin{enumerate}
\item \textbf{A per-language evaluation of decision models.} We report results for 27 European languages, 11 sources and 9 task groups, and reach across 210 languages. To our knowledge this is the first such evaluation for any decision model.
\item \textbf{Calibration under source shift.} Miscalibration depends on the task, in the same direction at all three sizes, with no detectable language effect on parallel items. Pooled calibration error understates per-source error by a factor of two.
\item \textbf{Automation limits.} The automatable share tracks chance-corrected skill and how well confidence ranks right answers above wrong ones. Thresholds chosen in-sample do not transfer between splits.
\item \textbf{Base-model effects.} On identical data, base choice outweighed parameter count, and reach outside the training languages came from the base.
\item \textbf{Two pitfalls for multi-source decision data,} each with a diagnostic others can run: label-word collision, and accuracy that rises under attack.
\item \textbf{A model family with a full comparison,} including suites where it trails Jev and Kev.
\end{enumerate}

Section~\ref{sec:related} reviews related work. Sections~\ref{sec:model} and~\ref{sec:data} describe the model and data. Section~\ref{sec:setup} sets out the experimental design, and Section~\ref{sec:results} reports results. Section~\ref{sec:analysis} analyses training choices, Section~\ref{sec:limits} discusses limitations and Section~\ref{sec:conclusion} concludes.

\section{Related Work}\label{sec:related}

\paragraph{Decision models.} TypeSafe's Jev~\cite{r1} introduced the typed request format of a state and named questions, trained with a method the company calls Reinforcement Learning for Calibrated Decisions. Its architecture, data and parameter count are undisclosed. Kev~\cite{r2} is an open-weight reproduction on Qwen bases, and its public evaluation suites and serving code underpin much of the work reported here. Kev pairs a pretrained language model with a pointer head. Its three smaller models train a LoRA adapter on a frozen base, while Kev-27B fine-tunes every weight of a post-trained Qwen release. Clef~\cite{r3} offers Jev-compatible open-weight models on the same Qwen bases. Zuse uses the adapter design at all three sizes and targets European languages.

\paragraph{Calibration and selective prediction.} Temperature scaling~\cite{r6} is the standard post-hoc method for calibrating neural classifiers, and the Brier score~\cite{r7} and expected calibration error~\cite{r8} are the usual measures. Reject-option classification goes back to Chow~\cite{r9}, and Geifman and El-Yaniv~\cite{r10} applied risk-coverage analysis to deep networks. Calibration degrades under distribution shift~\cite{r37}, and multilingual models are known to be less calibrated outside high-resource languages~\cite{r38}. Our automation measure is an in-sample risk-coverage point, and we report its limits in Section~\ref{sec:metrics}.

\paragraph{Multilingual evaluation.} Belebele~\cite{r12} and SIB-200~\cite{r13} test reading comprehension and topic classification across many languages. ScandEval~\cite{r23} provides Scandinavian benchmarks, including the acceptability data we use. We use these sets to measure transfer, and we add a comparison of language-model bases, including European-developed ones.

\paragraph{Prompt injection.} Text inside a document can steer a language model~\cite{r11,r39}. Decision models are exposed to the same risk because documents are untrusted input, yet their fixed outputs make the effect simple to measure as a change in the selected option. We use that measure in Section~\ref{sec:injection}. Language models choosing among options are also sensitive to option order~\cite{r41}, and models can reach high scores through shortcuts in how a task is built~\cite{r40}.

\section{Model}\label{sec:model}

\subsection{Architecture}

\begin{figure*}[t]
\centering
\includegraphics[width=0.92\textwidth]{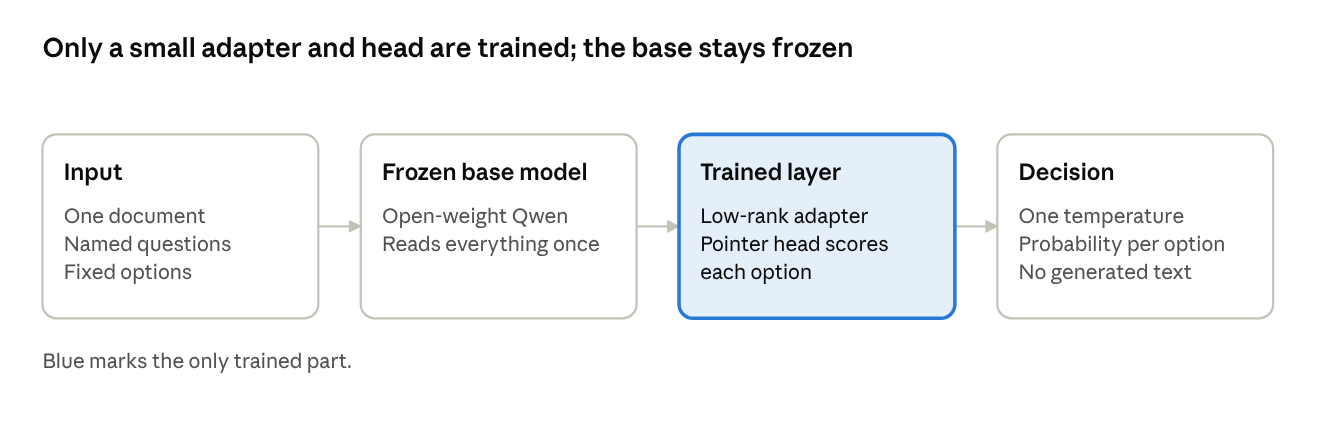}
\caption{Zuse architecture: document and question in, one probability per option out. The low-rank adapter sits inside the frozen base's layers.}\label{fig:arch}
\end{figure*}

Zuse uses a frozen open-weight language model as a reader (Figure~\ref{fig:arch}). Each question forms one token sequence that holds the state, the instruction and the options, following the layout used by Kev~\cite{r2}. Caller text is escaped so that it cannot contain the structural tokens. A pointer head reads the hidden states at the decision position and at each option, and gives each of the $K$ options a score $z_k$. Probabilities follow from a single temperature $T$~\cite{r6}:
\begin{equation}
p_k = \frac{\exp(z_k/T)}{\sum_{j=1}^{K}\exp(z_j/T)}
\end{equation}
Yes/no questions are two-option choices. Score questions use ordered options, and the expected position gives the score.

In most evaluations, each row carries one question. In serving, the state is encoded once and cached so that several questions can share it. This reuses Kev's serving stack~\cite{r2}, which we did not modify in design.

\subsection{Model family}

Only a low-rank LoRA adapter~\cite{r4} and the head are trained, with the base weights frozen. These two parts turn a text-generating language model into a decision model: the adapter changes how the base reads a request, and the head replaces text output with a probability over the supplied options. Table~\ref{tab:family} lists the family. All three bases are published under Apache-2.0~\cite{r27}.

\begin{table*}[t]
\centering
\caption{The Zuse model family. Memory is resident memory when served.}\label{tab:family}
\begin{tabular}{llcc}
\toprule
Model & Base & Memory, 16-bit & Weights, 4-bit\\
\midrule
Zuse-27B & Qwen3.8-27B & 65.5 GB & 15.1 GB\\
Zuse-9B & Qwen3.5-9B-Base & 21.9 GB & 5.6 GB\\
Zuse-4B & Qwen3.5-4B-Base & 14.3 GB & 3.1 GB\\
\bottomrule
\end{tabular}
\end{table*}

\subsection{Training}

The adapter and head are trained with a cross-entropy objective and a permutation-consistency term, which penalises disagreement between a row and a copy of it with shuffled options. Augmentations include none-of-the-above and injection. In none-of-the-above augmentation, that option appears as both the correct and an incorrect answer, following Kev~\cite{r2}. In injection augmentation, a planted instruction in the state names an option while the label stays unchanged, and that named option is sometimes correct (Section~\ref{sec:injection}). The temperature is fitted on a development set by minimising negative log-likelihood. We do not disclose loss weights, schedules or other training settings.

\section{Training Data}\label{sec:data}

We do not disclose the sources, composition or size of the training data, which we treat as commercially sensitive. The models target 27 European languages. No customer data was used in training, and no output of a commercial API model was used. Section~\ref{sec:limits} sets out what this non-disclosure means for readers.

\section{Experimental Setup}\label{sec:setup}

\subsection{Evaluation suites}\label{sec:suites}

Table~\ref{tab:suites} lists the suites and how each was used. The status column matters for interpretation, because several suites influenced development.

\begin{table*}[t]
\centering
\caption{Evaluation suites and their use during development.}\label{tab:suites}
\begin{tabularx}{\textwidth}{@{}p{3.1cm}Yp{3.3cm}Y@{}}
\toprule
Suite & Content & Size & Use during development\\
\midrule
European suite & 11 sources excluded from training, 9 task groups, 27 languages & 20,044 rows & Model selection, error diagnosis, design of later runs\\
Transfer v4 and v9~\cite{r2} & Public English suites & 764 rows in Transfer v4 per split & Development split used for selection and data choices; test split scored on five 27B checkpoints\\
SemIf~\cite{r2} & Authored English decisions plus perturbations & 144 and 108 rows & Monitored during development\\
Breadth report~\cite{r2} & 14 public datasets in 5 areas & 1,990 records, development split & Scored after freezing\\
When2Call~\cite{r30} & Tool-use choice & 3,423 tasks & Scored after freezing\\
World languages & Belebele~\cite{r12} and SIB-200~\cite{r13} & 210 languages, 32,700 rows & Scored after freezing; European rows overlap the European suite's sources\\
Long documents & Synthetic documents, 2k to 32k tokens & 1,350 rows & Scored on synthetic documents outside training\\
\bottomrule
\end{tabularx}
\end{table*}

The European suite contains Belebele reading~\cite{r12}, SIB-200 topics~\cite{r13}, multilingual MMLU~\cite{r36,r14}, two Norwegian knowledge sets~\cite{r32}, a sentiment set~\cite{r35}, Swiss court judgments~\cite{r24}, MInDS-14 banking intents~\cite{r21}, ScaLA acceptability in four Nordic languages~\cite{r23}, a German retrieval set~\cite{r34} and French yes/no questions~\cite{r33}. None of its sources appears in training, but some of its tasks resemble tasks seen in training. The suite therefore tests new sources rather than wholly new tasks. The German retrieval set is near ceiling for all three models (99.5 to 100\%).

Two sources in Transfer v4 need caution, because data used in training or in earlier runs may share origin with them. We did not confirm row-level overlap. Without those two sources, Zuse-27B's accuracy changes by about one point or less on both splits.

\subsection{Metrics}\label{sec:metrics}

For $N$ decisions with confidence $c_i$, the highest option probability, we report accuracy, the multi-class Brier score~\cite{r7} and expected calibration error over 15 equal-width bins~\cite{r8}.

The \textbf{automatable share} at error budget $b$ is the largest fraction of decisions, taken in descending order of confidence, whose error rate stays within $b$:
\begin{equation}
A(b) = \max\Big\{\frac{n}{N} : \frac{1}{n}\sum_{i=1}^{n} \mathbf{1}[\hat{y}_{(i)} \neq y_{(i)}] \le b \Big\}
\end{equation}
This follows the risk-coverage view of selective prediction~\cite{r10}. The threshold is chosen on the rows it is scored on, and per-group figures use a separate threshold for each group, so $A(b)$ is an in-sample upper bound. Thresholds do not transfer cleanly: a threshold set on the Transfer v4 test split gives 8.9\% error against a 5\% budget on the development split, and in the reverse direction 2.4\% error at 61\% coverage. We write \autofive{} for $A(0.05)$.

\textbf{Order flips} are the share of choice rows whose answer changes when options are reordered. We report this against reversed order and, separately, against a third random order, with and without averaging two orders. Both are measured on 1,223 choice rows from a 1,500-row sample of the European suite.

\textbf{Planted-instruction following} is the share of the same 1,500 rows on which the model selects an option named by a planted instruction. Its baseline is how often the clean model picks that option unprompted. The planted text has one wording in each of ten languages, including English, uses the English wording for all other languages, and always names a wrong option.

\subsection{Selection procedure and statistical treatment}\label{sec:selection}

Almost all configurations were trained once. We selected models using the European suite together with the Transfer v4 development split, and used both to diagnose errors and to plan later runs. The English test splits were scored on five 27B checkpoints. Their accuracies span 0.859 to 0.866, and test scores were recorded alongside flagship decisions, so independence from selection cannot be claimed.

On a 764-row split, an accuracy near 0.86 has a 95\% margin of about 2.5 points. For the European suite the margin is about 0.5 points under an independence assumption, and wider in practice because rows cluster within sources. Retraining on identical data moved two small-model results by up to 0.5 points, which is our only direct measure of training noise. Seeds for those retrainings are unconfirmed, and source-level scores vary far more than 0.5 points.

\section{Results}\label{sec:results}

Unless stated otherwise, scores come from the final models at 16-bit precision. The products available today are 4-bit builds, whose results we report separately. The breadth report and When2Call were run against deployed 4-bit endpoints.

\subsection{English public suites}

Table~\ref{tab:english} gives results on the public suites, with Jev and Kev figures as reported by the Kev project~\cite{r2}. Jev has only been run on development splits, so test and development results cannot be compared across systems. On the Transfer v4 development split, Zuse-27B is 2.5 points below Jev. An unpaired test on that accuracy gap gives $p = 0.19$, but paired predictions are not available, so this is only indicative. Jev is also ahead on Brier score and \autofive{}. Kev-27B is 2.9 points ahead of Zuse-27B on the test split. On the same bases, Kev-9B and Kev-4B lead Zuse-9B and Zuse-4B by 4.8 to 6.3 points on Transfer v4, with lower Brier scores. Zuse-27B scores 0.924 on SemIf against Kev-27B's reported 0.903, but SemIf is close to saturated and the Kev project does not use it to rank models.

\begin{table*}[t]
\centering
\caption{English public suites. Cells show accuracy / Brier / \autofive{} where available; Kev cells show accuracy / Brier. Jev and Kev figures are as reported by Kev~\cite{r2}.}\label{tab:english}
\footnotesize
\setlength{\tabcolsep}{4pt}
\resizebox{\linewidth}{!}{\begin{tabular}{@{}llllllll@{}}
\toprule
Suite and split & Zuse-27B & Zuse-9B & Zuse-4B & Jev & Kev-27B & Kev-9B & Kev-4B\\
\midrule
Transfer v4, development & 0.832 / 0.246 / 0.576 & 0.772 / 0.316 / 0.410 & 0.755 / 0.342 / 0.380 & 0.857 / 0.211 / 0.70 & 0.851 / 0.225 & 0.820 / 0.289 & 0.817 / 0.269\\
Transfer v4, test & 0.860 / 0.204 / 0.776 & 0.797 / 0.279 / 0.598 & 0.775 / 0.313 / 0.512 & -- & 0.889 / 0.156 & 0.852 / 0.217 & 0.838 / 0.242\\
Transfer v9, development & 0.786 / 0.305 / 0.462 & 0.716 / 0.382 / 0.302 & 0.695 / 0.410 / 0.305 & -- & -- & -- & --\\
Transfer v9, test & 0.787 / 0.285 / 0.609 & 0.723 / 0.369 / 0.485 & 0.694 / 0.415 / 0.392 & -- & -- & -- & --\\
SemIf, 144 authored rows & 0.924 & 0.847 & 0.882 & 0.965 & 0.903 & -- & --\\
\bottomrule
\end{tabular}}
\end{table*}

\subsection{Breadth and tool choice}

Zuse-27B is behind Jev on the breadth report, and not clearly separated from Kev-27B. Table~\ref{tab:breadth} shows a chance-corrected index from 0 to 100, built from 14 public datasets. The difference from Jev is 5.8 points (95\% paired bootstrap CI 2.2 to 9.0). The difference from Kev-27B is 2.7 points (95\% CI $-0.2$ to 5.3).

The Jev and Kev rows were measured by the Kev project~\cite{r2}. The Jev rows were collected in September 2026 through a gateway, and the model revision is not exposed. Zuse rows are our own, and were run against a deployed 4-bit endpoint rather than the 16-bit evaluation build. That run met server errors, and all 1,990 records were answered after retries. This is a report in the style of a community index, not a third-party score.

\begin{table}[t]
\centering
\caption{Breadth report on the development split, by area.}\label{tab:breadth}
\small
\resizebox{\linewidth}{!}{\begin{tabular}{@{}lcccc@{}}
\toprule
Area & Jev & Kev-27B & Z-27B & Z-9B\\
\midrule
Knowledge, reasoning & 37.7 & 33.8 & 36.0 & 24.6\\
Language understanding & 77.4 & 75.6 & 69.7 & 57.8\\
Retrieval, classification & 75.4 & 62.9 & 63.0 & 57.7\\
Tools, automation & 63.6 & 64.2 & 59.0 & 58.5\\
Arts, human taste & 12.7 & 14.7 & 10.0 & 6.7\\
\midrule
\textbf{Overall} & \textbf{53.3} & \textbf{50.2} & \textbf{47.5} & \textbf{41.1}\\
95\% CI lower & 50.7 & 47.4 & 44.8 & 38.4\\
95\% CI upper & 56.3 & 53.3 & 50.7 & 44.2\\
\bottomrule
\end{tabular}}
\end{table}

On When2Call~\cite{r30}, a four-way choice (chance 25\%) between using a tool, answering directly, asking for more information and saying it cannot help, and using our own instruction wording, Zuse-27B scored 57.5\% and Zuse-9B 39.7\% over 3,423 tasks. Mean confidence was 72.9\% and 67.8\%, so both models were overconfident on this task.

\subsection{European suite}

Table~\ref{tab:european} gives the results on the European suite. The pooled calibration error is low, but this masks differences between groups. Within sources it ranges from 0.007 to 0.091, reaching 0.091 on the Norwegian quiz, 0.082 on grammar and 0.079 on multilingual MMLU. By language it reaches 0.067, largely through task mix. Across the 14 breadth datasets, on the deployed endpoint and with 10 bins, the mean is 0.100. Weighted by rows, the per-source error is 0.035, more than twice the pooled figure, because overconfidence on some sources offsets underconfidence on others. Calibration should therefore be checked for any specific task.

This miscalibration is structured by task rather than language. Belebele and SIB-200 give parallel items in 26 of the 27 languages, 400 rows each. On those items the gap between mean confidence and accuracy has a standard deviation across languages of 0.015 for Zuse-27B, and 0.014 and 0.013 for the smaller models, close to the sampling noise of about 0.014 at this size. Across sources the gap ranges from $-0.071$ to $+0.091$ for Zuse-27B and up to $+0.144$ for Zuse-4B. The direction holds at all three sizes: multilingual MMLU and the Norwegian quiz are overconfident, and grammar is underconfident. The single temperature fitted on development data reduced error on 7 of the 11 sources for Zuse-27B, and temperatures fitted per source would range from about 0.25 to 2.5. A deployment that moves to a new task therefore needs its own calibration, while moving to a new language on a known task needs less.

\begin{table*}[t]
\centering
\caption{European suite. Order flips use 1,223 choice rows and planted instructions a 1,500-row sample.}\label{tab:european}
\resizebox{\linewidth}{!}{\begin{tabular}{@{}lccc@{}}
\toprule
Measure & Zuse-27B & Zuse-9B & Zuse-4B\\
\midrule
Accuracy (\%) & \textbf{87.5} & 84.1 & 81.2\\
Brier score & 0.185 & 0.229 & 0.264\\
Expected calibration error & 0.016 & 0.023 & 0.032\\
Automatable at 1 / 2 / 5 / 10\% error (\%) & 40 / 58 / 77 / 94 & 30 / 50 / 67 / 86 & 26 / 42 / 61 / 78\\
Order flips, reversed order (\%) & 5.6 & 6.2 & 8.9\\
Order flips against a third order, single / two averaged (\%) & 3.7 / 3.3 & 4.4 / 3.5 & 6.9 / 5.2\\
Planted instruction followed, baseline in brackets (\%) & 8.1 (5.8) & 12.2 (7.1) & 14.2 (8.3)\\
Accuracy at 4-bit (\%) & 87.4 & 83.6 & 80.6\\
Brier score at 4-bit & 0.188 & 0.236 & 0.275\\
Expected calibration error at 4-bit & 0.016 & 0.026 & 0.034\\
\autofive{} at 4-bit (\%) & 76.0 & 66.3 & 58.6\\
Separate long-document set, 2k to 32k tokens (\%) & 98.1 & 98.1 & 97.6\\
\bottomrule
\end{tabular}}
\end{table*}

\begin{figure*}[t]
\centering
\includegraphics[width=0.74\textwidth]{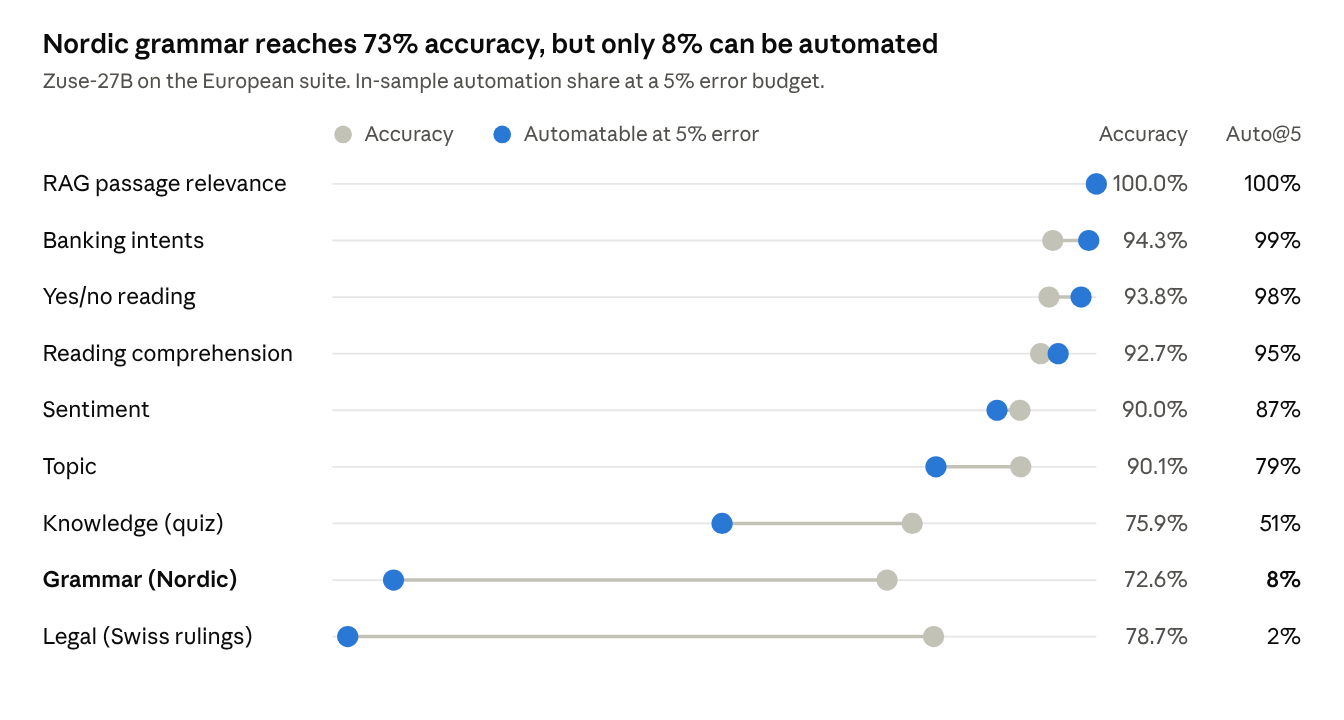}
\caption{Accuracy and in-sample automation share by task group, Zuse-27B.}\label{fig:groups}
\end{figure*}

Figure~\ref{fig:groups} shows the automatable share by task group. It mostly follows accuracy (Spearman 0.93 across the nine groups), but accuracy is not corrected for chance. Nordic grammar is a clear case: 72.6\% on balanced labels, with only 8\% automatable at a 5\% error budget. On Swiss court rulings Zuse-27B reaches 78.7\%, below the 79.3\% from always answering no, so it shows no gain over the majority answer, and only 1.5\% can be automated. Banking intents and reading comprehension reach far higher shares. Across the 11 sources and three models, the automatable share correlates more strongly with chance-corrected skill, taking the larger of the majority-class and uniform baselines, than with raw accuracy (Spearman 0.97 against 0.91 over 33 pairs). It correlates most strongly with how well confidence ranks right answers above wrong ones (AUROC, Spearman 0.98 over the 31 pairs where it is defined). These are in-sample upper bounds with per-group thresholds (Section~\ref{sec:metrics}).

\begin{figure*}[t]
\centering
\includegraphics[width=0.66\textwidth]{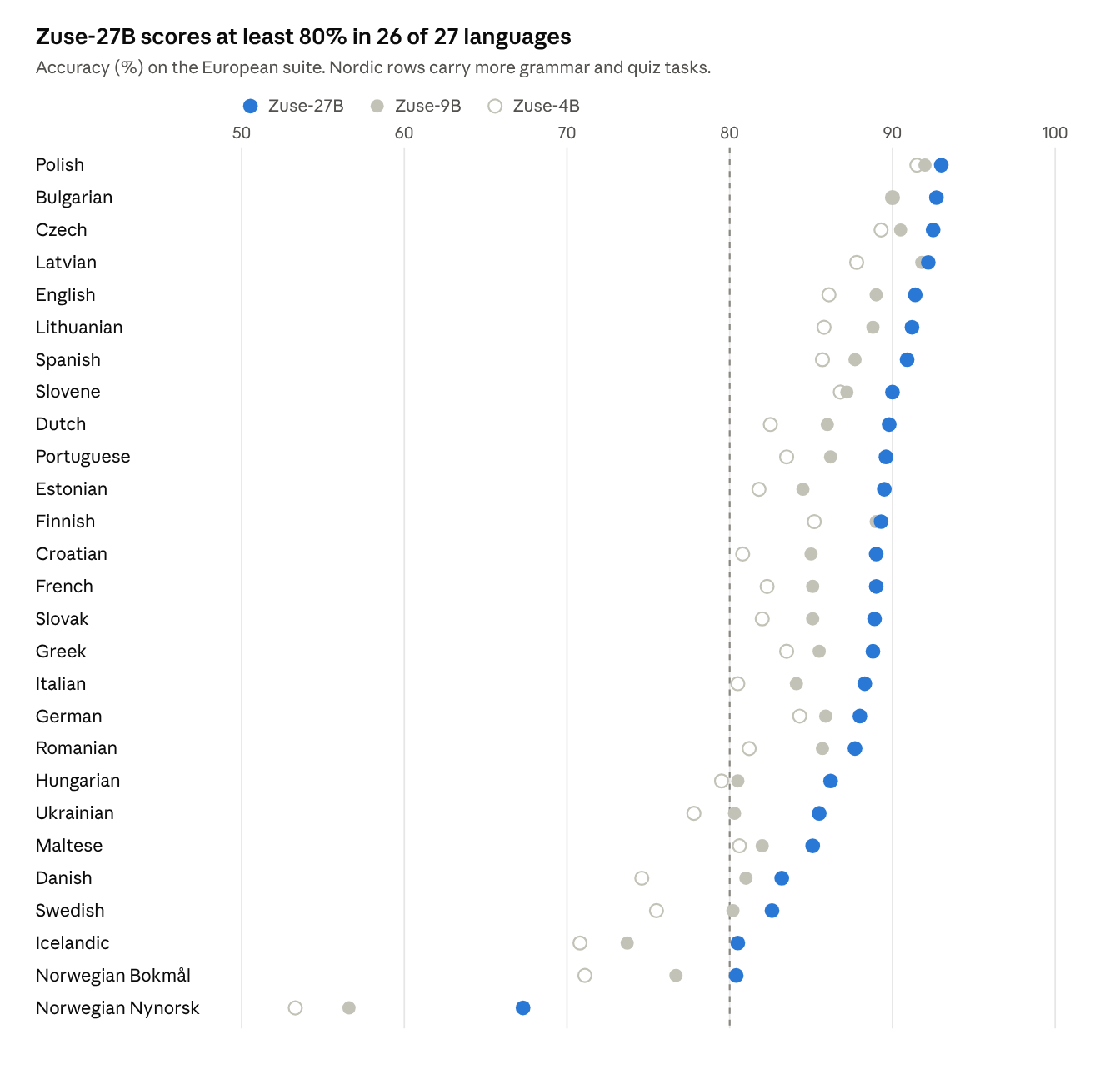}
\caption{Accuracy by language on the European suite, all three models.}\label{fig:langs}
\end{figure*}

Figure~\ref{fig:langs} shows results by language. Icelandic (600 rows) and Bokm\aa{}l (1,400 rows) sit within sampling noise of 80\%. Language averages reflect task mix as well as language, since Nordic languages carry the grammar and quiz sources. On Belebele reading and SIB-200 topics in the final model, those languages score between 87 and 96.

\subsection{World languages}

Table~\ref{tab:world} reports reach across languages, including the languages trained on. Reading covers 122 languages from Belebele, and topic classification 205 from SIB-200, for 210 languages and 32,700 rows in total. We grouped languages using an approximate list of 140 codes derived by hand from the language list Qwen published for Qwen3, applied here to later Qwen versions.

\begin{table*}[t]
\centering
\caption{Accuracy (\%) on world languages, by whether the language is in the Qwen list. Cells show reading / topic.}\label{tab:world}
\begin{tabular}{@{}lccccc@{}}
\toprule
Group & Reading languages & Topic languages & Zuse-27B & Zuse-9B & Zuse-4B\\
\midrule
In the list & 97 & 139 & 85.1 / 87.8 & 76.5 / 86.1 & 71.0 / 84.6\\
Outside the list & 25 & 66 & 60.5 / 65.2 & 48.7 / 59.6 & 44.8 / 56.2\\
\bottomrule
\end{tabular}
\end{table*}

On reading and topic together, Zuse-27B reaches 85\% or above in 115 languages and falls below 60\% in 25. Four of the five weakest are written in N'Ko, Ol Chiki or Tifinagh, at 28 to 31\%, and the weakest, Central Kanuri in Arabic script, scores 22\%. These have topic rows only, where uniform chance is 14.3\% and always choosing the largest class gives 22 to 26\%.

\subsection{Inference speed}

Table~\ref{tab:speed} reports model time at 16-bit precision on one B300 GPU, using Kev's serving stack. On 100 records, this path differed from the unoptimised path by at most 0.019 in probability, with at most one changed answer per model. In a 4-bit deployment on a 24 GB card, a two-question request took about 230 ms on a cached document. On 24 GB cards, documents are limited to 8,192 tokens.

\begin{table*}[t]
\centering
\caption{Model time per request, new document / cached document, and throughput.}\label{tab:speed}
\resizebox{\linewidth}{!}{\begin{tabular}{@{}lcccc@{}}
\toprule
Model & 2 questions (ms) & 6 questions (ms) & 5 questions, 2,200 tokens (ms) & Requests per second, 1 / 32 clients\\
\midrule
Zuse-27B & 32 / 18 & 45 / 31 & 182 / 51 & 20 / 47\\
Zuse-9B & 13 / 8 & 17 / 12 & 86 / 20 & 49 / 118\\
Zuse-4B & 10 / 6.5 & 14 / 10 & 84 / 18 & 62 / 141\\
\bottomrule
\end{tabular}}
\end{table*}

\section{Analysis of Training Choices}\label{sec:analysis}

This section compares single-seed runs, so it is observational rather than a controlled study. Runs in a comparison may also differ in settings we do not report, so these comparisons are indicative. We treat gaps under 0.5 points on the European suite as noise (Section~\ref{sec:selection}).

\subsection{Base model and size}

The largest effect we observed came from the base model. On an early corpus and identical data, moving from Qwen3.5-4B to Qwen3.8-27B raised accuracy on held-out sources from 75.4 to 85.4. On the same data and recipe, Qwen3.8-27B scored 85.6 against 75.8 for EuroLLM-22B~\cite{r28}, and led in all 27 languages. Borealis 2 31B~\cite{r29}, a Norwegian-trained model on a Gemma base, reached 82.9. On the same data, EuroLLM-22B also scored below Qwen3.5-4B (77.8) and Qwen3.5-9B (80.0), and on the early corpus Mistral-Nemo-12B scored 66.1 against 75.4 for Qwen3.5-4B, so base choice outweighed parameter count in these runs. Borealis 2 did not lead Qwen3.8-27B even on Bokm\aa{}l (76.0 against 77.1) and was at chance on grammar (49.5). These comparisons have limits: the recipe was developed on Qwen bases, and the bases differ in post-training.

Within each size, world-language scores varied little across the nine checkpoints we scored, staying within 1.2 points of that size's final model for languages in the Qwen list and within 2.1 outside it. Between sizes they differ by 14 to 16 points on reading. This suggests that reach outside the training languages comes largely from the base. The evidence is limited, because all nine checkpoints trained on similar data and there is no control.

\subsection{Label-word collision}\label{sec:collision}

Adding one topic-labelled data family lowered accuracy on the SIB-200 topic test by about 10 points. The closest isolated pair of runs differs only in that family: 87.6 without it and 77.3 with it. Overall accuracy on the European suite moved from 86.7 to 83.4.

The cause appears to be shared label words. The added family used two label words that also name SIB-200 categories, with narrower meanings. On the affected run, the model predicted one of those categories 2.4 times as often as it occurs. Datasets merged for training may therefore conflict through their label vocabulary even when their languages and domains look compatible.

\subsection{Injection training and shortcut learning}\label{sec:injection}

Our first injection augmentation produced results that looked strong and proved misleading. On a Qwen3.5-4B run, following of planted instructions fell from 37.6\% to 0.2\%, but accuracy under attack rose above clean accuracy, from 75.6 to 81.7. Every planted note had named a wrong option, so the model could learn that a named option is incorrect.

In a run that also dropped the reordered copies, we changed the augmentation so that planted notes sometimes name the correct option. The headline figure then looked worse, at 16.8\% following against a 9.6\% baseline, with accuracy under attack at 71.9 against 75.5 clean. This is the more honest result, because the model is no longer exploiting the construction of the attack. In the final Zuse-27B, following is 8.1\% against a 5.8\% baseline, and accuracy falls from 87.1 to 85.6 on the same sample. Robustness evaluations should therefore check for accuracy that improves under attack.

\subsection{Order stability}

Adding reordered copies of rows had almost no effect on order flips (13.6\% to 13.1\%, in a step that also added injection training), whereas a permutation-consistency loss helped. Across 27B runs, flips against a third order averaged 6.4\% in twelve runs without the loss and 4.5\% in eight runs with it. Those groups also differ in data, so the comparison is indicative. At a stronger setting on the 4B model, the loss cut flips from 10.5\% to 6.8\% and cost about 1.5 accuracy points. At a weaker setting on the 27B model we saw no accuracy cost.

\section{Limitations and Threats to Validity}\label{sec:limits}

\paragraph{Model limitations.} Knowledge-heavy questions remain weak in this training recipe: 75.9\% on the quiz group, and 0.675 on MMLU-Pro~\cite{r31} (200 development rows), equal to Kev-27B's reported 0.675 and below Jev's reported 0.84. Nynorsk scores 67.3\%. Order sensitivity and instruction following are reduced but not removed. Accepted input runs to 64k tokens on larger GPUs, but accuracy above 32k is unmeasured because we lacked enough held-out documents. There is no image input.

\paragraph{Selection bias.} The European suite shaped model selection, error diagnosis and the design of later runs, including removal of one data family and one targeted run. The Transfer v4 development split shaped data choices and flagship selection. Scores on both are likely to be optimistic. The English test splits were seen several times, and their scores were recorded alongside flagship decisions.

\paragraph{Missing comparisons.} We have not run Jev, Kev or Clef on the European suite, so we make no claim about relative European performance. Nor have we compared frontier language models with structured output on the same rows.

\paragraph{Comparability.} Jev and Kev figures come from the Kev project. Jev's revision is unknown and its automation figure uses a development split. Zuse's breadth scores came from a deployed endpoint rather than the evaluation build used elsewhere.

\paragraph{Statistical power.} Results come from single seeds, and our noise estimate rests on two small-model retraining pairs. English test splits are small. Rows within the European suite are clustered by source, so simple margins understate uncertainty.

\paragraph{Construct validity.} Not all training labels were independently checked. Long-document rows are simple fact checks. The planted-instruction test uses few wordings and always names a wrong option. All results come from public benchmarks or synthetic documents, not from a live deployment.

\paragraph{Undisclosed training details.} We do not disclose the training data or the full training recipe, so the training cannot be reproduced from this paper, and readers must rely on our statement that the evaluation sources in the European suite were excluded from training. We compared the source identifiers in the final training list with the suite's 11 sources and found no match, but did not compare row text for that final list. Public suites can be run against the 4-bit Zuse-27B and Zuse-9B. Zuse-4B, the 16-bit builds and the sampled European suite rows are not available.

\section{Conclusion}\label{sec:conclusion}

We presented Zuse, a family of adapter-based decision models targeting 27 European languages, and reported its results across English, European and world-language evaluations. On European sources excluded from training the largest model reaches 87.5\% accuracy, though that figure is a development result. On English public suites it is behind Jev and Kev on every aggregate measure where both were scored. Four observations are likely to carry beyond this system, though each rests on single-seed runs of one model family: miscalibration follows the task more than the language, so calibration should be checked per task; the automatable share follows chance-corrected skill; base-model choice outweighed parameter count and set language reach; and merged datasets can conflict through their label vocabulary, while robustness results should be checked for shortcuts.

The priority for future work is a clean evaluation: a new European test suite, scored once and alongside open decision models. Repeated seeds for the main comparisons would resolve several of the confounds identified here.

\section*{Availability}

Readers can test Zuse-27B through a live demo at \url{https://bineric.com/zuse}, which needs no API key, or through the Bineric API. Zuse-27B and Zuse-9B can also be run in a reader's own cloud account as 4-bit models through AWS Marketplace, and an Azure Marketplace offer is in preparation. Deployment guides and sample inputs are at \url{https://github.com/binericai/models/tree/main/zuse}. Zuse weights are not openly released, and Zuse-4B is not offered. The training data and training code are not released.

\section*{Competing Interests}

The author is the founder of Bineric AI, which sells Zuse, and an assistant professor at OsloMet. This work was carried out at Bineric AI and was not part of the author's duties at OsloMet. AI tools assisted with drafting and auditing this paper.

\section*{Acknowledgements}

This work was completed in part at the Open Models Codefest, part of the NVIDIA, Deutsche Telekom, and Open Hackathons Program, using NVIDIA B300 GPUs. The author would like to acknowledge NVIDIA for their support. Zuse builds on Qwen models by the Qwen team at Alibaba Cloud under the Apache License 2.0, and on Kev by Jared Palmer under the Apache License 2.0.

\end{document}